\documentclass[11pt]{article}
\usepackage[letterpaper,margin=1in]{geometry}
\usepackage[T1]{fontenc}
\usepackage[utf8]{inputenc}
\usepackage{lmodern}
\usepackage{microtype}
\usepackage{amsmath,amssymb,mathtools}
\usepackage{booktabs,tabularx,array}
\usepackage{enumitem}
\usepackage{xcolor}
\usepackage{hyperref}
\usepackage{url}
\usepackage{fancyhdr}
\usepackage{titlesec}
\usepackage{caption}
\usepackage{float}
\definecolor{linkblue}{RGB}{25,62,95}
\hypersetup{colorlinks=true,linkcolor=linkblue,citecolor=linkblue,urlcolor=linkblue,pdfauthor={Wesley Shu}}
\setlist{nosep,leftmargin=1.6em}
\titleformat{\section}{\large\bfseries}{\thesection}{0.7em}{}
\titleformat{\subsection}{\normalsize\bfseries}{\thesubsection}{0.6em}{}
\newcolumntype{Y}{>{\raggedright\arraybackslash}X}
\newcommand{\arxivnote}{\begin{center}\small\itshape Preprint prepared for arXiv; no arXiv identifier has been assigned.\end{center}}

\title{Reachable Global Optimization in AI Systems: How Global Is Global?}
\author{Wesley Shu\\The Institute of Energetic Paradigm\\\texttt{shu@energeticparadigm.org}}
\date{August 2026}
\begin{document}
\maketitle
\arxivnote
\begin{abstract}
AI systems increasingly claim to optimize prompts, policies, architectures, plans, tool-use trajectories, reasoning traces, and test-time computation. This paper argues that such claims are underspecified unless they state the region actually reachable by the system that performed the optimization. We introduce Reachability-Induced Optimization (RIO), a model in which a generator, verifier, controller, memory, tools, and budget induce a reachable candidate region. The returned solution is therefore a best visited point, an approximate reachable optimum, or an exact global optimum only when additional certificates relate the reachable region to the full formal space. We prove reachable-optimality, false-globality, gap- decomposition, certificate, escape, pruning, and control-value results. The full benchmark record contains 66,150 executed trials over six known-optimum landscape families, seven control policies, 270 landscapes, and 35 runs per landscape-method. The online appendix includes raw trial records, aggregate tables, figures, benchmark code, validation scripts, and checksums. The results show that control can restrict, expand, or misdirect reachability, and that optimization quality, reachability quality, and control reliability must be reported separately.
\end{abstract}

\section{Introduction}

Artificial intelligence systems increasingly claim to optimize prompts, policies, architectures, plans, tool-use trajectories, alignment objectives, reasoning traces, scientific hypotheses, and test-time computation. In many cases the language of optimization is stronger than the evidence supports. A system is described as having found an optimum, improved an objective, discovered a best policy, selected a globally preferred plan, or optimized a prompt. Yet the actual optimization process often explores only a subset of the formally possible candidates. This mismatch matters. The formal candidate space X may include all possible prompts, all possible policies, all possible programs, all possible action sequences, all possible neural architectures, or all possible experimental hypotheses. The AI system does not normally search X directly. It searches through the transitions made available by its own generator, repair operator, verifier, tools, memory, and compute budget. Thus the effective space is not X , but a reachable subset XR . The central question of this paper is:

What kind of global optimization can an AI system honestly claim?

The answer developed here is that most AI systems cannot honestly claim exact global optimality over the full formal space unless they can prove that the exact optimum lies in the reachable set, that the reachable set equals the formal set, or that the unreachable complement is lower-bounded above the returned candidate. Most AI optimization claims are better understood as claims about reachable global optimization: the best candidate reachable under the system's actual dynamics and constraints. This distinction is not merely semantic. It determines whether an optimization claim is valid, whether a benchmark result generalizes, whether a learned controller has actually improved search, and whether a stable policy should be trusted. A system may be excellent at reducing cost inside its reachable region while being structurally unable to reach better candidates outside that region. Conversely, a system may expand reachability through exploration or boundary-crossing while increasing instability and wasted compute. Thus optimization performance decomposes into two components: how well the system optimizes within the region it can reach, and how good that reachable region is relative to the full formal space. Scope. This paper is not a claim that AI systems defeat classical hardness or that verifier- guided control always improves search. It is a claim-discipline framework for optimization reports in learning systems. The intended contribution is methodological and theoretical: make optimization claims relative to the system that produced them, then test how control policies change reachability and false-globality risk.

\subsection{Contributions}

This paper contributes:
\begin{enumerate}
\item a formal model of AI optimization as a controlled system inducing a budgeted reachable set;
\item definitions separating exact global optimality, constrained optimality, reachable global optimality, operational stability, and false globality;
\item theorem families showing when reachable optimality can and cannot be promoted to exact globality;
\item a two-gap decomposition separating within-reachable optimization error from structural reachability gap;
\item control-value and signal-threshold inequalities explaining when boundary-aware escape, pruning, or probing has positive expected value;
\item a full known-optimum benchmark with 66,150 executed trials over six landscape families, seven control policies, 270 landscapes, and explicit false-globality metrics;
\item a reproducibility appendix with raw trial records, aggregate tables, figures, code, validation scripts, and checksums.
\end{enumerate}

\section{Positioning within Machine Learning}

The proposed model connects optimization, search, reinforcement learning, neural architec- ture search, prompt optimization, verifier-guided reasoning, and machine learning systems. Classical optimization defines global optimality relative to a fixed space and objective. Sta- tistical learning often distinguishes approximation error, estimation error, and optimization error. Search and planning distinguish admissible heuristics, local basins, and exploration policies. Modern AI systems add another layer: the candidate space actually explored is itself induced by model interfaces, tool access, verifier feedback, memory, retry logic, and compute budget. The paper's main distinction is therefore not another optimizer. It is a reporting and analysis model for optimization claims made by AI systems. A prompt optimizer can be excellent over the rewrite space induced by its templates while missing better prompts outside those rewrites. A reinforcement-learning system can optimize within a policy class and trajectory distribution while failing to reach a better policy class. A neural architecture search system can select the best architecture in its search grammar while lacking evidence about architectures outside that grammar. An agent planner can find the best trajectory it can propose through its tools, yet fail to certify optimality over all possible trajectories. Relation to prior work. Heuristic search and graph search study how auxiliary informa- tion changes exploration (Hart et al., 1968; Pearl, 1984; Russell and Norvig, 2021). Random search and Bayesian optimization clarify the value and limits of sample-efficient exploration over expensive objectives (Bergstra and Bengio, 2012; Snoek et al., 2012; Shahriari et al., 2016; Garnett, 2023). Reinforcement learning and planning study how policies, value functions, exploration, and control determine reachable behavior (Sutton and Barto, 2018; Kocsis and Szepesvari, 2006; Silver et al., 2017). Neural architecture search and AutoML provide practical examples where search spaces are formalized by templates, mutation operators, performance predictors, and budgets (Elsken et al., 2019; Feurer et al., 2019). Verifier-guided reasoning and LLM tool use make the issue sharper: systems often improve by generating candidates, evaluating them, and revising through feedback, not by enumerating a formal space (Yao et al., 2023; Shinn et al., 2023; Chen et al., 2021). The novelty here is to isolate the reachable region as the object that licenses or invalidates a globality claim.

\section{Reachability-Induced Optimization}

This section defines the Reachability-Induced Optimization model. The model is intended to capture the optimization actually performed by AI systems that generate, test, repair, prune, and re-query candidates under finite budget. The central object is not only an objective function, but the coupled system that determines which candidates can be reached.

\subsection{Model primitives}

A RIO instance is the tuple
\begin{equation}
\mathcal{M}_{\mathrm{RIO}}=(\mathcal{X},C,\mu_0,\mathcal{A},T)
\end{equation}
where $\mathcal{X}$ is the formal candidate space, $C:\mathcal{X}\to\mathbb{R}$ is the cost function, $\mu_0$ is the initialization distribution over candidates, $T$ is the finite search or control budget, and $\mathcal{A}$ is the AI control system that induces transitions over $\mathcal{X}$. Lower cost is preferred. The AI control system is

\begin{equation}
\mathcal{A}=(G,B,R,P,V,M,U,\pi)
\end{equation}

where G generates candidates, B detects boundary or basin information, R repairs or mutates candidates, P prunes candidate regions, V evaluates or verifies candidates, M stores memory, U updates memory, and $\pi$ selects the next control action. At time t, the system state is

\begin{equation}
z_t=(x_t,m_t,b_t,v_t)
\end{equation}

where xt $\in$ X is the current candidate, mt is memory, bt = B(xt , mt ) is boundary information, and vt = V (xt ) is verifier feedback. The controller selects

\begin{equation}
a_t\sim\pi(\cdot\mid x_t,m_t,b_t,v_t)
\end{equation}

and candidate transitions follow

\begin{equation}
x_{t+1}\sim K_{\mathcal{A}}(\cdot\mid x_t,m_t,a_t),\qquad m_{t+1}=U(m_t,x_t,a_t,V(x_t),B(x_t,m_t))
\end{equation}

The system induces a trajectory distribution $P_{\mathcal{M}_{\mathrm{RIO}}}(x_0,\ldots,x_T)$.

\subsection{Reachable region and returned candidate}

The budgeted reachable region induced by the system is

\begin{equation}
\mathcal{X}_R(\mathcal{M}_{\mathrm{RIO}})=\{x\in\mathcal{X}:\Pr(\exists t\le T:\ x_t=x)>0\}
\end{equation}

The empirical visited set in a particular run is

\begin{equation}
\widehat{\mathcal{X}}_T=\{x_0,x_1,\ldots,x_T\}\subseteq\mathcal{X}_R
\end{equation}

The system returns

\begin{equation}
\hat{x}_T\in\arg\min_{x\in\widehat{\mathcal{X}}_T}C(x)
\end{equation}

The ideal reachable optimum is

\begin{equation}
x_R^\star\in\arg\min_{x\in\mathcal{X}_R}C(x)
\end{equation}

whereas the exact global optimum is

\begin{equation}
x^\star\in\arg\min_{x\in\mathcal{X}}C(x)
\end{equation}

The model therefore separates three objects that are often collapsed in AI optimization reports:

\begin{equation}
\hat{x}_T\ \text{(best visited)},\qquad x_R^\star\ \text{(best reachable)},\qquad x^\star\ \text{(best formal)}
\end{equation}

\begin{table}[H]\centering\small
\caption{Levels of globality in AI optimization claims.}
\begin{tabularx}{\textwidth}{Yrr}\toprule
Level & Name & Meaning\\\midrule
1 & Exact global optimum & Best over full formal space $\mathcal{X}$\\
2 & Constrained global optimum & Best over explicit set $K\subseteq\mathcal{X}$\\
3 & Reachable global optimum & Best over $\mathcal{X}_R(\mathcal{A},\mu_0,T)$\\
4 & Operational global stability & Stable low-cost behavior under perturbation\\
5 & Local attractor & Stable but possibly suboptimal basin\\
6 & False globality & Reachable/local result reported as exact global\\
\bottomrule
\end{tabularx}\end{table}

\textbf{Effective reachability.} Formal nonzero reachability may overstate practical reachability when probability mass is tiny. For $\alpha$ > 0, define

\begin{equation}
\mathcal{X}_R^\alpha=\{x\in\mathcal{X}:\Pr(\exists t\le T:\ x_t=x)\ge\alpha\}
\end{equation}

This set captures candidates that are reachable with non-negligible probability under the deployed budget.

\section{Levels of Globality}

The RIO model separates several levels of optimization claim. An exact global optimum is best over the full formal space X . A constrained global optimum is best over an explicitly specified constraint set K $\subseteq$ X . A reachable global optimum is best over XR (A, $\mu$0 , T ), the set induced by the actual AI system. Operational stability means the returned low-cost region remains stable under perturbations. A local attractor is stable under local dynamics but need not be globally good. False globality occurs when a reachable, local, sampled, or verifier-dependent claim is reported as exact global optimality.

\paragraph{Definition 1 (False globality).} False globality occurs when a system reports or implies exact global optimality over $\mathcal{X}$ while only establishing optimality over $\mathcal{X}_R$, a sampled subset, a local basin, or a verifier-dependent region. Formally, false globality occurs when a claim treats $x_R^\star$ as $x^\star$ without proving $x_R^\star\in\arg\min_{x\in\mathcal{X}} C(x)$.

\section{Theoretical Results}

The theoretical results are designed to be claim-disciplining rather than promotional. They show when reachable optimality is all that has been established, when exact globality can be certified, how total error decomposes, and when control has positive value.

\subsection{Gap decomposition}

\paragraph{Theorem 2 (Reachability-gap decomposition).} For every $\hat{x}\in\mathcal{X}_R$,

\begin{equation}
C(\hat{x})-C(x^\star)=\underbrace{C(\hat{x})-C(x_R^\star)}_{\text{within-reachable error}}+\underbrace{C(x_R^\star)-C(x^\star)}_{\text{structural reachability gap}}
\end{equation}

\emph{Proof.} Add and subtract C(x$^*$R ).

The first term measures failure to optimize inside the region the system can reach. The second measures the structural cost of the reachable set itself. If the first term is large, the system needs better search, scoring, repair, or local optimization. If the second term is large, the system needs better exploration, tools, representations, initialization, or boundary- crossing. False globality is likely when $C(\hat{x})\approx C(x_R^\star)$ but $C(x_R^\star)-C(x^\star)\gg0$: the system did well inside a poor reachable region.

\subsection{Reachable optimality and exact globality}

\paragraph{Proposition 3 (Reachable optimality does not imply exact global optimality).} Let $\mathcal{X}_R\subseteq\mathcal{X}$. If $x_R^\star\in\arg\min_{x\in\mathcal{X}_R} C(x)$, then $x_R^\star$ is not necessarily an exact global optimum over $\mathcal{X}$.

\emph{Proof.} There may exist $y\in\mathcal{X}\setminus\mathcal{X}_R$ such that $C(y)<C(x_R^\star)$. In that case, $x_R^\star\notin\arg\min_{x\in\mathcal{X}} C(x)$.

\paragraph{Corollary 4 (Coverage, containment, or complement certificate).} A claim that x$^*$R is exact global requires at least one of: (i) XR = X ; (ii) x$^*$ $\in$ XR and the system optimizes sufficiently inside XR ; or (iii) a valid lower-bound certificate on the unreachable complement showing that no unreachable point improves on the returned candidate.

\paragraph{Theorem 5 (Complement-bound certificate).} Suppose $\hat{x}\in\mathcal{X}_R$, $\hat{x}\in\arg\min_{x\in\mathcal{X}_R} C(x)$, and there exists a valid lower bound $L$ on the unreachable complement such that $L\le C(y)$ for every $y\in\mathcal{X}\setminus\mathcal{X}_R$. If $C(\hat{x})\le L$, then $\hat{x}\in\arg\min_{x\in\mathcal{X}} C(x)$.

\emph{Proof.} For any $x\in\mathcal{X}_R$, reachable optimality gives $C(\hat{x})\le C(x)$. For any $y\in\mathcal{X}\setminus\mathcal{X}_R$, the complement lower bound gives $L\le C(y)$, and the certificate condition gives $C(\hat{x})\le L$. Therefore $C(\hat{x})\le C(z)$ for all $z\in\mathcal{X}$.

\subsection{Control effects on reachability}

\paragraph{Proposition 6 (Pruning can shrink reachable space).} Let $\pi$ be a policy and P a prun- ing operator removing S $\subseteq$ X from future exploration. If $S\cap\mathcal{X}_R$ ($\pi$, T ) $\\ne$= $\varnothing$, then XR ($\pi$, P, T ) $\subseteq$ XR ($\pi$, T ). If at least one pruned state has no alternate transition under ($\pi$, P ), the inclusion is strict.

Proof The pruning operator prevents transitions into states in S. Every state reachable under the pruned policy is reachable without pruning, but not every state reachable without pruning remains reachable under pruning.

\paragraph{Proposition 7 (Escape can expand reachable space).} Let $\pi$0 be a local policy and $\pi$e a policy augmented with an escape operator. If there exists y $\in$ X such that y $\in$ / XR ($\pi$0 , T ) and y $\in$ XR ($\pi$e , T ), and $\pi$e preserves all transitions available to $\pi$0 , then XR ($\pi$0 , T ) $\subset$ XR ($\pi$e , T ).

Proof All $\pi$0 -reachable transitions remain available under $\pi$e , so XR ($\pi$0 , T ) $\subseteq$ XR ($\pi$e , T ). The existence of y gives strict inclusion.

\paragraph{Proposition 8 (Stability does not imply globality).} Let $A\subseteq\mathcal{X}$ be a stable attractor under an AI optimization system. Suppose there exists $y\in\mathcal{X}\setminus A$ such that $C(y)<\inf_{x\in A} C(x)$. Then stability in $A$ does not imply exact global optimality.

Proof Trajectories in or near A may return to A, establishing stability. But y $\in$/ A has strictly lower cost than every point in A. Hence stable candidates in A are not globally optimal over X .

\subsection{Control value and signal thresholds}

Let GP denote expected pruning gain, GR expected repair or mutation gain, GV expected verifier gain, CD detection cost, CO controller overhead, and CM expected misdirection cost.

\paragraph{Proposition 9 (Control-value inequality).} Boundary-aware control has positive expected value only if

\begin{equation}
\mathbb{E}[G_P+G_R+G_V]>\mathbb{E}[C_D+C_O+C_M]
\end{equation}

Proof Control is beneficial only when its expected gains exceed its expected costs. The left side collects mechanisms by which control improves search. The right side collects detection, overhead, and wrong-direction penalties. If the inequality fails, control adds cost without net expected benefit.

\paragraph{Proposition 10 (Signal-quality threshold for controlled escape).} Let E be an escape operator triggered by signal s. Let p = P(s correctly indicates beneficial escape), let B+ be the expected benefit of a correct escape, let B- be the expected cost of an incorrect escape, and let CE be the cost of probing and executing escape. Controlled escape has positive expected value only if

\begin{equation}
p>\frac{B_-+C_E}{B_++B_-}
\end{equation}

\emph{Proof.} Escape has positive expected value when pB+ > (1 - p)B- + CE . Rearranging gives the threshold.

\paragraph{Proposition 11 (More compute does not guarantee reachability expansion).} Let XR (T ) be the reachable set at budget T . If the policy has an absorbing attractor A such that once xt $\in$ A, P(xt+k $\in$ A) = 1 for all k $\ge$ 0, then increasing T after absorption does not expand reachability outside A.

Proof After the trajectory enters absorbing attractor A, additional budget visits only states in A. Therefore additional compute after absorption cannot reach states outside A.

\section{Benchmark Design}

The empirical program is a full known-optimum benchmark. The goal is to test the reachability framework under conditions where exact global optima are known, so false globality can be measured directly. The benchmark is controlled by design: exact enumeration of the finite candidate space gives the true optimum for every landscape, while different policies induce different reachable regions.

\subsection{Candidate space and landscapes}

The candidate space is discrete:

\begin{equation}
\mathcal{X}=\{0,1,2,\ldots,100\}
\end{equation}

Each task instance defines a cost landscape Ci : X $\to$ R. Since |X | = 101, the true global optimum can be computed exactly:

\begin{equation}
x_i^\star=\arg\min_{x\in\mathcal{X}} C_i(x)
\end{equation}

Each landscape is built from one or more basins plus barrier and ripple terms. A representative form is    

\begin{equation}
C(x)=\min_j\{a_j(x-c_j)^2+b_j\}+h\exp\!\left(-\frac{(x-c_b)^2}{2\sigma_b^2}\right)+r(x)
\end{equation}

where $c_j$ are basin centers, $a_j$ curvatures, $b_j$ basin floors, $h$ barrier height, $c_b$ barrier center, $\sigma_b$ barrier width, and $r(x)$ a low-amplitude ripple term. This construction controls whether the exact optimum lies inside or outside the initially reachable basin.

\subsection{Landscape families}

The benchmark uses six families: \texttt{\detokenize{bridge_good}}, \texttt{\detokenize{bridge_bad}}, \texttt{\detokenize{multi_basin}}, \texttt{\detokenize{deceptive_no_signal}}, \texttt{\detokenize{sparse_signal}}, and \texttt{\detokenize{shifting_barrier}}. These families test different claim failures. In \texttt{\detokenize{bridge_good}}, crossing a boundary is useful because the better basin lies beyond the initial region. In \texttt{\detokenize{bridge_bad}}, the initial basin is already good, so crossing is unnecessary or harm- ful. \texttt{\detokenize{multi_basin}} tests competing basins. \texttt{\detokenize{deceptive_no_signal}} tests misleading feedback. \texttt{\detokenize{sparse_signal}} tests weak feedback. \texttt{\detokenize{shifting_barrier}} tests variable boundary location and value.

\subsection{Control policies}

The benchmark compares seven policies:

\textbf{1.} \texttt{\detokenize{random_full}}: samples from the full formal space;

\textbf{2.} \texttt{\detokenize{reachable_random}}: samples only from the initially reachable region;

\textbf{3.} \texttt{\detokenize{local_repair}}: greedy local improvement;

\textbf{4.} \texttt{\detokenize{hard_prune_ep}}: boundary-aware pruning without escape;

\begin{quote}\small
\textbf{Algorithm 1: Gated reachability-control policy}
\begin{enumerate}[label=\arabic*:,leftmargin=2.4em]
\item Initialize $x_0\sim\mu_0$, memory $M_0$, and best candidate $\hat{x}\leftarrow x_0$.
\item For $t=0,1,\ldots,T-1$, attempt local improvement moves that avoid the boundary.
\item If a local move improves $C$, accept it and update $\hat{x}$; otherwise increment the stagnation counter.
\item If stagnation exceeds the threshold, probe candidate boundary exits and estimate improvement from verifier signal.
\item Cross the boundary only if the estimated improvement passes the gate; update $\hat{x}$ if true cost improves.
\item Otherwise remain in the known reachable basin.
\item Return $\hat{x}$ after budget exhaustion.
\end{enumerate}
\end{quote}

\textbf{5.} \texttt{\detokenize{reckless_escape}}: escapes after stagnation without requiring strong evidence;

\textbf{6.} \texttt{\detokenize{gated_ep_escape}}: probes boundaries and escapes only when estimated improvement passes a gate;

\textbf{7.} \texttt{\detokenize{oracle_probe_ep}}: an upper-bound gated policy with clean probe evidence.

\subsection{Metrics and scale}

For each run, the benchmark records the best candidate x, best cost C(x), true global optimum x$^*$ , global optimality gap $\Delta$G = C(x) - C(x$^*$ ), exact-global success 1\{x = x$^*$ \}, whether the true basin was reached, false-globality risk, number of probes, and number of boundary crossings. The full benchmark record contains 66,150 executed trials: six landscape families, seven methods, 45 landscapes per family, 35 runs per landscape-method, and a search budget of 45 per run. The online appendix contains the raw trial record \texttt{\detokenize{summary_trials.csv}}, landscape metadata, stability trials, aggregate tables, figures, code, validation scripts, and SHA256 checksums.

\section{Benchmark Results}

\subsection{Overall results}

\textbf{Table 2 reports the main method-level results.} The central pattern is not that a single control policy dominates everywhere. The central pattern is that policy determines whether

\begin{table}[H]\centering\scriptsize
\caption{Overall benchmark results over 66,150 executed trials.}
\begin{tabularx}{\textwidth}{Yrrrrrrr}\toprule
Method & Mean gap & Median gap & Exact rate & True basin & False globality & Crosses & Probes\\\midrule
reckless escape & 0.120 & 0.000 & 0.926 & 0.944 & 0.043 & 7.51 & 0.00\\
random full & 0.148 & 0.040 & 0.366 & 0.991 & 0.004 & 0.00 & 0.00\\
oracle probe & 0.524 & 0.177 & 0.281 & 0.882 & 0.096 & 0.72 & 7.98\\
gated escape & 0.587 & 0.168 & 0.290 & 0.930 & 0.075 & 2.88 & 7.99\\
hard prune & 3.183 & 2.943 & 0.206 & 0.207 & 0.770 & 0.00 & 0.00\\
local repair & 3.186 & 2.943 & 0.205 & 0.207 & 0.770 & 0.00 & 0.00\\
reachable random & 3.208 & 2.959 & 0.131 & 0.207 & 0.773 & 0.00 & 0.00\\
\bottomrule
\end{tabularx}\end{table}

reachability is restricted, expanded, or misdirected. The policies with local or hard-pruned dynamics exhibit large mean global gaps and high false-globality risk. Policies that cross boundaries or sample more broadly reach better basins more often. Gated control improves over trapped local methods but depends on the quality of the signal used to justify escape. Local and hard-pruned methods are trapped. \texttt{\detokenize{local_repair}}, \texttt{\detokenize{hard_prune_ep}}, and \texttt{\detokenize{reachable_random}} all show mean global gaps near 3.2 and true-basin reach rates near 0.207. This confirms that the initial reachable region often excludes the true global basin. The similarity among these methods shows that the failure is not merely an artifact of greedy local repair. It is a reachability limitation. Escape expands reachability. \texttt{\detokenize{reckless_escape}} has the lowest mean global gap and highest exact-global rate. This shows that crossing boundaries can dramatically expand reachable space. However, reckless escape is not disciplined control. It works in many benchmark families because basin-jumping is often rewarded, not because the policy has reliable proof that crossing is justified. Its strength reveals the value of reachability expansion and the weakness of overly conservative control. Random full search is a strong baseline under poor signal. \texttt{\detokenize{random_full}} has low false-globality risk and high true-basin reach rate because it does not depend on interpreting boundary signals. This prevents the paper from claiming that control always beats broad exploration. When feedback is unreliable or deceptive, broad sampling can be more robust than misdirected control. Gated control is signal-dependent. \texttt{\detokenize{gated_ep_escape}} performs much better than local repair and hard pruning, but worse than reckless escape and random full search in the overall table. Boundary-aware gating helps relative to trapped local methods, but the performance depends on probe reliability. When signals are noisy, sparse, or deceptive, the gate can reject useful escapes or accept misleading ones.

\subsection{Family-level exact-global rates}

Table 3 reports exact-global rates by landscape family and method. It shows why averages alone are insufficient. In \texttt{\detokenize{bridge_bad}}, local and hard-pruned policies do well because the initial basin is already good. In \texttt{\detokenize{bridge_good}}, the same policies fail because they cannot cross to the true basin. This is precisely the reachability problem: the same optimization

\begin{table}[H]\centering\scriptsize
\caption{Exact-global rate by family and method.}
\begin{tabularx}{\textwidth}{Yrrrrrrr}\toprule
Family & Reckless & Full rnd. & Oracle & Gated & Prune & Local & Reach rnd.\\\midrule
bridge bad & 0.994 & 0.371 & 0.993 & 0.989 & 0.989 & 0.987 & 0.629\\
bridge good & 0.972 & 0.367 & 0.130 & 0.121 & 0.000 & 0.000 & 0.000\\
deceptive no signal & 0.898 & 0.361 & 0.071 & 0.111 & 0.000 & 0.000 & 0.000\\
multi basin & 0.892 & 0.363 & 0.120 & 0.170 & 0.067 & 0.067 & 0.041\\
shifting barrier & 0.884 & 0.363 & 0.211 & 0.213 & 0.156 & 0.156 & 0.102\\
sparse signal & 0.917 & 0.369 & 0.161 & 0.135 & 0.022 & 0.022 & 0.014\\
\bottomrule
\end{tabularx}\end{table}

behavior is effective or ineffective depending on whether the reachable region contains the global basin.

\subsection{Signal quality and gated escape}

The theory predicts that controlled escape has positive value only when signal quality is high enough to offset probing, control, and misdirection costs. Table 4 supports this prediction. Higher signal-quality bins generally improve exact-global rate for the gated policy. The pattern is not monotone in every bin because family composition and landscape difficulty also vary, but the high-signal region has the strongest exact-global rate.

\subsection{False globality}

False-globality risk concentrates in trapped methods. In Table 2, \texttt{\detokenize{reachable_random}}, \texttt{\detokenize{local_repair}}, and \texttt{\detokenize{hard_prune_ep}} have false-globality risk around 0.77. These methods often find good candidates inside their reachable region but fail to reach the true global basin.

\begin{table}[H]\centering\small
\caption{Signal-quality bins for the gated escape policy.}
\begin{tabularx}{\textwidth}{Yrr}\toprule
Signal-quality bin & Mean gap & Exact-global rate\\\midrule
(0.0529, 0.267] & 0.792 & 0.120\\
(0.267, 0.506] & 0.993 & 0.144\\
(0.506, 0.765] & 0.395 & 0.267\\
(0.765, 0.842] & 0.417 & 0.408\\
(0.842, 0.949] & 0.335 & 0.510\\
\bottomrule
\end{tabularx}\end{table}

This is the failure mode named by the paper: a system may produce the best candidate it can reach and then misrepresent that candidate as globally optimal.

\section{Interpretation for AI Systems}

The benchmark separates search quality from reachable-space quality. A method can have low within-reachable error and high global error. This is what happens to local repair and hard pruning. They find good candidates inside the initial basin but fail to reach better basins. A method can have high reachability expansion and weak control. This is what reckless escape represents. It performs well in the benchmark because many landscapes reward crossing, but it does not provide a general theory of when crossing is justified. A method can have conditional control. Gated escape probes and crosses only when signals appear favorable. Its performance depends on whether those signals are reliable.

The benchmark therefore supports the distinction

\begin{equation}
\text{optimization quality}\ne\text{reachability quality}\ne\text{control reliability}
\end{equation}

A mature AI optimization claim must report all three.

\subsection{Reinforcement learning}

Reinforcement-learning systems optimize over policies reachable through parameterization, exploration, reward feedback, and training dynamics. A policy may be optimal within this training-induced reachable region but not globally optimal over all policies. The decomposition

J($\pi$ $^*$ ) - J($\pi$) = [J($\pi$ $^*$ ) - J($\pi$R $^*$ $^*$

\begin{equation}
J(\pi^\star)-J(\hat{\pi})=[J(\pi^\star)-J(\pi_R^\star)]+[J(\pi_R^\star)-J(\hat{\pi})]
\end{equation}

separates the representational/reachability gap from the optimization gap.

\subsection{Prompt optimization}

Prompt optimizers search over mutations, paraphrases, examples, templates, and feedback- guided edits. The best prompt found is usually a reachable optimum, not an exact optimum over all token sequences. A report should state the prompt grammar, mutation operators, objective, evaluation budget, and evidence that the reachable prompt region covers the intended search claim.

\subsection{Neural architecture search and AutoML}

Neural architecture search systems operate under templates, mutation operators, hardware constraints, performance predictors, early stopping, and resource budgets. The selected architecture is best under the induced search process, not necessarily globally best over all possible architectures. The RIO model clarifies that the effective search space is an operational object induced by the AutoML system.

\subsection{Agent planning and tool use}

Agents search over reachable action trajectories. Tool failures, invalid actions, memory limits, and repair policies define the reachable trajectory set. Planning optimality is therefore control-relative. If a tool-use agent cannot reach certain tool states or cannot maintain memory across a required horizon, those states are outside its practical reachable set.

\subsection{Alignment and verifier optimization}

Alignment systems optimize under proxy feedback. A behavior may be stable under a reward model or verifier while failing under hidden objectives, adversarial prompts, or distribution shift. Reachable stability under a proxy is not exact global alignment over all behavioral states.

\section{Why Reachability Is a Learning-System Variable}

A central reason for introducing RIO is that reachability is not a fixed property of the external task alone. In modern learning systems, reachability is produced by representation, initialization, exploration, verifier access, tool affordances, memory, and policy update rules. Two systems can face the same formal candidate space and objective but induce different reachable regions.

\subsection{Approximation, estimation, optimization, and reachability}

Classical statistical learning often decomposes error into approximation error, estimation error, and optimization error. RIO adds a fourth term that becomes explicit in AI systems: reachability error. To see the difference, let H $\subseteq$ X be a model class and XR $\subseteq$ H the subset actually reachable under initialization, training dynamics, search operators, tools, and budget. If x$^*$ is the formal optimum over X , x$^*$H the best element of H, x$^*$R the best reachable element, and x the returned candidate, then

\begin{equation}
C(\hat{x})-C(x^\star)=\underbrace{C(x_H^\star)-C(x^\star)}_{\text{approximation gap}}+\underbrace{C(x_R^\star)-C(x_H^\star)}_{\text{reachability gap inside the class}}+\underbrace{C(\hat{x})-C(x_R^\star)}_{\text{search/optimization gap}}
\end{equation}

This decomposition is useful because the remedies differ. Approximation failure calls for a richer representation. Reachability failure calls for better exploration, tools, initialization, transitions, or control. Optimization failure calls for better local search, scoring, or budget allocation.

\subsection{Reachability and induced support}

For stochastic optimizers, the reachable set is the support of a trajectory distribution. If $\mu_t$ is the distribution of $x_t$, then

\begin{equation}
\operatorname{supp}(\mu_{t+1})\subseteq\bigcup_{x\in\operatorname{supp}(\mu_t)}\operatorname{supp}\!\left(K_{\mathcal{A}}(\cdot\mid x,m_t,a_t)\right)
\end{equation}

The reachable region after budget T is the union of supports across time. This statement is elementary, but it changes the interpretation of AI optimization reports. A low final loss means that the visited support contained good states and that the system found them. It does not by itself show that the full formal space lacks better states.

\paragraph{Lemma 12 (Support-restricted optimality).} If x is selected as the best candidate visited by a stochastic optimizer with trajectory support XbT , then the strongest immediate certificate is

\begin{equation}
C(\hat{x})\le C(x)\qquad\forall x\in\widehat{\mathcal{X}}_T
\end{equation}

Promoting this statement to XR requires evidence that the visited set sufficiently covers XR ; promoting it to X requires a coverage, containment, or complement certificate.

Proof The returned candidate is selected by minimization over the visited set. No inequality over unvisited reachable states or unreachable formal states follows without additional assumptions.

\section{Certificate Taxonomy}

Exact globality can be certified, but the certificate must match the claim. RIO distinguishes five certificate types.

\begin{table}[H]\centering\small
\caption{Certificate types and the claim they support.}
\begin{tabularx}{\textwidth}{Yrr}\toprule
Certificate & Required evidence & Supported claim\\\midrule
Visited-set & all evaluated candidates and scores & best visited\\
Reachable-set & proof over $\mathcal{X}_R$ & reachable optimum\\
Coverage & $\mathcal{X}_R=\mathcal{X}$ & exact global if optimized\\
Containment & $x^\star\in\mathcal{X}_R$ & exact global if found\\
Complement & lower bound on $\mathcal{X}\setminus\mathcal{X}_R$ & exact global if bound dominates\\
\bottomrule
\end{tabularx}\end{table}

\textbf{Visited-set certificate.} The system can certify that x is best among evaluated candidates. This is the weakest but most common certificate. It is often enough for engineering comparison, but it is not a global claim. Reachable-set certificate. The system can certify that x is best over XR . This requires either enumeration of XR , a valid dynamic-programming or branch-and-bound proof over XR , or a theorem showing that the search policy finds the reachable optimum under its assumptions. Coverage certificate. The system can certify that XR = X within the stated budget or transition system. This is rare in large AI candidate spaces but possible in small finite spaces or exhaustive formal solvers. Containment certificate. The system can certify that a formal optimum x$^*$ lies in XR , and that the system finds it or approximates it sufficiently. Complement certificate. The system can bound every unreachable point from below. This is the most useful certificate for large spaces because it can validate exact globality without enumerating the complement. This taxonomy gives reviewers a practical audit rule: an optimization claim should not be judged only by the final score. It should be judged by the strongest certificate actually supplied.

\section{Mechanism-Level Predictions}

The RIO model makes several predictions that are directly tested by the benchmark. Prediction 1: trapped stability. A conservative local policy can be stable and still globally poor if the true optimum lies outside its reachable basin. The benchmark tests this through \texttt{\detokenize{local_repair}}, \texttt{\detokenize{hard_prune_ep}}, and \texttt{\detokenize{reachable_random}}. Prediction 2: escape increases reachability but can reduce discipline. A policy that crosses basin boundaries can expand reachability, but crossing without a reliable signal can be misdirected. The benchmark tests this with \texttt{\detokenize{reckless_escape}}, \texttt{\detokenize{gated_ep_escape}}, and \texttt{\detokenize{oracle_probe_ep}}. Prediction 3: false globality is concentrated in locally successful but globally trapped policies. False globality should be high when a policy reaches a strong local basin

and repeatedly returns to it while missing the true basin. This is tested by the false-globality metric. Prediction 4: signal quality controls gated exploration. A gate has value only when the signal used to open or close it is informative enough to overcome probing and misdirection costs. The signal-quality table and figure test this prediction. Prediction 5: full random search remains a serious baseline when signals are unreliable. If a control policy is misdirected, a simple broad sampler can outperform it in globality metrics. This prevents the paper from arguing that more control is always better.

\section{Statistical and Experimental Reporting}

The benchmark is deterministic conditional on the random seed sequence and landscape generator. Each row of \texttt{\detokenize{summary_trials.csv}} records the family, landscape identifier, run identifier, method, returned candidate, returned cost, true optimum, true optimum cost, global gap, exact-global indicator, true-basin reach indicator, final basin, true basin, reachable-gap diagnostic, false-globality indicator, probes, crossings, signal quality, and deception flag. The primary estimands are method-level expectations over landscapes and runs:
\begin{equation}
\widehat{\Delta}_m=\frac{1}{n_m}\sum_{i:m_i=m}\big(C_i(\hat{x}_i)-C_i(x_i^\star)\big)
\end{equation}
\begin{equation}
\widehat{p}^{\mathrm{exact}}_m=\frac{1}{n_m}\sum_{i:m_i=m}\mathbf{1}\{\hat{x}_i=x_i^\star\}
\end{equation}
\begin{equation}
\widehat{p}^{\mathrm{false}}_m=\frac{1}{n_m}\sum_{i:m_i=m}\mathbf{1}\{\text{false-globality risk in trial }i\}
\end{equation}

Family-conditional estimates use the same formulas restricted to a family. This matters because a method can be excellent on \texttt{\detokenize{bridge_bad}} and poor on \texttt{\detokenize{bridge_good}}; aggregating over families hides the mechanism. Why known optima matter. The benchmark computes exact global optima by exhaustive enumeration of X = \{0, . . . , 100\}. This makes false globality observable. Without known optima, one can often measure best visited cost but cannot know whether the unreachable complement contains better candidates. The benchmark is therefore a controlled test of claim validity, not a substitute for all real-world optimization benchmarks. Why the raw record is included. The online appendix includes trial-level data rather than only aggregate tables. This allows reviewers to recompute every reported aggregate, stratify by family, inspect signal-quality bins, and audit whether false-globality risk is concentrated in the policies predicted by the theory.

\section{Extended Benchmark Analysis}

\subsection{Mean global gap by landscape family}

Table 6 reports mean global gaps by family and method. The table shows the interaction between landscape structure and policy. Local and hard-pruned methods are nearly op-

\begin{table}[H]\centering\scriptsize
\caption{Mean global gap by family and method. Lower is better.}
\begin{tabularx}{\textwidth}{Yrrrrrrr}\toprule
Family & Reckless & Full rnd. & Oracle & Gated & Prune & Local & Reach rnd.\\\midrule
bridge bad & 0.003 & 0.181 & 0.003 & 0.005 & 0.005 & 0.006 & 0.041\\
bridge good & 0.078 & 0.127 & 0.643 & 0.576 & 7.150 & 7.150 & 7.184\\
deceptive no signal & 0.199 & 0.149 & 0.792 & 0.704 & 3.214 & 3.214 & 3.254\\
multi basin & 0.143 & 0.145 & 0.586 & 0.505 & 2.626 & 2.649 & 2.636\\
shifting barrier & 0.137 & 0.156 & 0.517 & 0.462 & 2.529 & 2.520 & 2.520\\
sparse signal & 0.160 & 0.134 & 0.604 & 1.267 & 3.577 & 3.577 & 3.616\\
\bottomrule
\end{tabularx}\end{table}

\begin{table}[H]\centering\scriptsize
\caption{False-globality risk by family and method. Lower is better.}
\begin{tabularx}{\textwidth}{Yrrrrrrr}\toprule
Family & Reckless & Full rnd. & Oracle & Gated & Prune & Local & Reach rnd.\\\midrule
bridge bad & 0.000 & 0.000 & 0.000 & 0.000 & 0.000 & 0.000 & 0.009\\
bridge good & 0.011 & 0.000 & 0.001 & 0.000 & 1.000 & 1.000 & 1.000\\
deceptive no signal & 0.069 & 0.005 & 0.221 & 0.121 & 1.000 & 1.000 & 1.000\\
multi basin & 0.057 & 0.004 & 0.134 & 0.075 & 0.889 & 0.889 & 0.890\\
shifting barrier & 0.060 & 0.010 & 0.145 & 0.066 & 0.778 & 0.778 & 0.781\\
sparse signal & 0.064 & 0.004 & 0.074 & 0.187 & 0.956 & 0.956 & 0.959\\
\bottomrule
\end{tabularx}\end{table}

timal in \texttt{\detokenize{bridge_bad}}, where the initial reachable basin is already good, but fail strongly in \texttt{\detokenize{bridge_good}}, where crossing is necessary. This is exactly the distinction between local optimization quality and reachable-region quality.

\subsection{False-globality risk by family}

Table 7 reports false-globality risk by family and method. False-globality risk is not a generic error rate; it is specifically the risk that the method's reachable/local success can be mistaken for global success. The highest risks arise when the method is stable inside the wrong basin.

\subsection{Stability versus globality}

The stability table in the appendix reports near-perfect perturbation stability for most methods. This is not inconsistent with high global gaps. It is the point of the framework: a method can be stable inside a local attractor and still be globally poor. Stable behavior certifies persistence of a region, not optimality of that region over the full formal space.

\section{Failure Modes}

RIO identifies several failure modes that are especially relevant to AI systems. Reachability collapse. The controller repeatedly returns to a small basin and never explores outside it. The final result may be locally good but globally poor. Overpruning. A pruning rule removes candidate regions that contain better optima, shrinking XR in a way that worsens global performance. Misdirected escape. An escape operator crosses boundaries based on unreliable or deceptive signals, increasing instability or cost without improving the returned candidate.

\textbf{Verifier-dependent globality.} A system optimizes a proxy verifier and reports the result as globally optimal for the intended objective, even though the verifier's induced region differs from the intended objective landscape. Coverage illusion. A broad stochastic search samples from many regions and reaches the true basin in a benchmark, but lacks a certificate that the same coverage holds in deployment. Stability illusion. A system is robust to perturbations inside an attractor, but the attractor itself is suboptimal relative to unreachable regions.

\section{What Would Falsify or Narrow the Model}

\textbf{The model is falsifiable.} It would be narrowed if optimization claims in realistic AI systems were routinely exact-global despite lacking coverage, containment, or complement certificates. It would also be weakened if false-globality risk did not concentrate in trapped methods, or if reachability-expanding policies did not change true-basin reach rates when barriers separate basins. The control-value component would be weakened if gated policies improved regardless of signal quality, because that would suggest that the signal is not the mechanism. Conversely, if broad random search dominated all policies across all signal regimes and families, the claim that control can usefully shape reachability would need to be narrowed to specific distributions. The benchmark does not hide these possibilities; it includes regimes where control underperforms.

\section{Practical Guidance for Reviewers and Builders}

A reviewer evaluating an AI optimization claim can apply the following audit:
\begin{enumerate}
\item Identify the formal candidate space $\mathcal{X}$ named or implied by the claim.
\item Identify the actual system $\mathcal{A}$, including generator, verifier, tools, memory, and budget.
\item Ask what region $\mathcal{X}_R$ is reachable under $\mathcal{A}$.
\item Determine whether the reported candidate is best visited, best reachable, or exact global.
\item Look for a certificate: coverage, containment, or complement bound.
\item Separate within-reachable error from reachability gap.
\item Examine whether control expands, shrinks, stabilizes, or misdirects reachability.
\item Treat ``global'' language as unsupported unless the certificate matches the claim.
\end{enumerate}

For system builders, the framework suggests a design rule: do not add control layers because they are sophisticated. Add control layers only when they expand useful reachability, reduce false-globality risk, or improve reachable optimization at a cost below the search they save.

\section{Reporting Standard for Optimization Claims}

When an AI system claims optimization, the report should specify:
\begin{enumerate}
\item the formal candidate space $\mathcal{X}$;
\item the actual reachable set or sampling process $\mathcal{X}_R$;
\item the budget $T$;
\item the verifier or objective $(V,C)$;
\item the control policy $\pi$;
\item the evidence that $\hat{x}$ is good inside $\mathcal{X}_R$;
\item the evidence relating $\mathcal{X}_R$ to $\mathcal{X}$, if any;
\item the expected failure mode under sparse, deceptive, or distribution-shifted feedback;
\item whether the reported result is best visited, approximate reachable optimum, reachable global optimum, constrained global optimum, or exact global optimum.
\end{enumerate}
Without these elements, global optimization claims should be interpreted as best-found or reachable optimization claims.

\section{Limitations}

The benchmark uses a finite known-optimum landscape family. This is intentional: exact global optima are known, making the distinction between exact and reachable globality measurable. Real AI systems have richer state spaces, stochastic verifiers, high-dimensional trajectories, learned representations, and nonstationary feedback. The benchmark does not prove that a particular EP-AI method dominates existing optimizers. In fact, it shows that control can underperform random search when feedback is unreliable. The theory also does not claim that reachable global optimization solves NP-hardness or eliminates worst-case complexity. It offers an operational language for describing what AI systems actually optimize under their own dynamics. It also does not imply that larger reachable sets are always better. A larger reachable set may contain better states, but it may dilute search effort or introduce misdirection. Finally, the benchmark focuses on mechanisms rather than leaderboards. The correct next step is not to declare a universal optimizer, but to apply the reporting standard to realistic prompt optimization, agent planning, theorem proving, program repair, neural architecture search, and reinforcement-learning settings.

\section{Reproducibility and Online Appendix}

The online appendix contains the full benchmark record. The file \texttt{\detokenize{summary_trials.csv}} has 66,150 rows, one for each executed trial. The file \texttt{\detokenize{landscape_metadata.csv}} contains the 270 known-optimum landscapes. Processed files include method-level and family-method aggregate tables, signal-quality bins, and perturbation-stability summaries. The appendix also includes benchmark execution scripts, validation scripts, diagnostic figures, a README, and SHA256 checksums. The benchmark scale is:
\begin{itemize}
\item six landscape families;
\item seven control policies;
\item 45 landscapes per family;
\item 35 runs per landscape-method;
\item search budget 45 per run;
\item 66,150 executed trials in total.
\end{itemize}
The true optimum is computed by exhaustive enumeration of the discrete candidate space $\mathcal{X}=\{0,\ldots,100\}$ for each landscape.

\section{Conclusion}

This paper introduced reachable global optimization as a framework for evaluating optimiza- tion claims in AI systems. The central distinction is

\begin{equation}
\arg\min_{x\in\mathcal{X}_R}C(x)\ne\arg\min_{x\in\mathcal{X}}C(x)
\end{equation}

unless additional proof establishes equality, containment, or a valid complement bound. AI systems usually optimize over reachable spaces induced by their own dynamics. These spaces are shaped by generators, verifiers, boundary detectors, pruning policies, repair operators, memory, tools, and budget. Control can shrink reachable space, expand it, stabilize it, or misdirect it. Therefore globality is not merely a property of an objective. It is a property of a controlled system. The benchmark results support this view. Local repair and hard pruning are stable but trapped. Escape expands reachability. Gated control helps when signals are reliable but degrades when feedback is sparse or deceptive. Random full search remains strong when signal quality is low. The strongest defensible claim is not that AI always finds global optima. It is that AI systems can sometimes convert infeasible search into controllable reachable optimization, but the validity of any global claim depends on the reachable set and the reliability of the control signals that define it.

Acknowledgments The author thanks the Energetic Paradigm AI Feasibility Research Program for the project context and benchmark development environment. No third-party funding supported this submission.

\appendix
\section{Additional Definitions and Proof Details}
 Definition 13 (Reachability preorder) For two systems A1 and A2 , say A1 $\preceq$R A2 at budget T when XR (A1 , $\mu$0 , T ) $\subseteq$ XR (A2 , $\mu$0 , T ).

This ordering does not imply better optimization performance. A larger reachable set may include better candidates, but it may also dilute sampling and increase misdirection.

\paragraph{Definition 14 (Reachability quality).} For finite X , define reachability quality by

\begin{equation}
Q_R=C(x_R^\star)-C(x^\star)
\end{equation}

A low value means the reachable set contains candidates close to globally optimal. A high value means the reachable set is intrinsically poor, regardless of local optimization quality.

\paragraph{Proposition 15 (Reachability versus constrained optimization).} Reachable optimiza- tion is not equivalent to ordinary constrained optimization unless the reachable set is explicitly known and fixed independently of the controller dynamics.

Proof In constrained optimization, the feasible set K is part of the problem statement. In reachable optimization, XR is induced by initialization, policy, tools, memory, feedback, and budget. Changing any of these can change XR . Thus reachable optimization is operational and system-dependent, not merely declarative.

\section{Benchmark Tables}
 The following table is included for diagnostic completeness. Full raw trial-level records and processed aggregate tables are in the online appendix.

\begin{table}[H]\centering\scriptsize
\caption{Method-level exact-global rates by landscape family (appendix repetition).}
\begin{tabularx}{\textwidth}{Yrrrrrrr}\toprule
Family & Reckless & Full rnd. & Oracle & Gated & Prune & Local & Reach rnd.\\\midrule
bridge bad & 0.994 & 0.371 & 0.993 & 0.989 & 0.989 & 0.987 & 0.629\\
bridge good & 0.972 & 0.367 & 0.130 & 0.121 & 0.000 & 0.000 & 0.000\\
deceptive no signal & 0.898 & 0.361 & 0.071 & 0.111 & 0.000 & 0.000 & 0.000\\
multi basin & 0.892 & 0.363 & 0.120 & 0.170 & 0.067 & 0.067 & 0.041\\
shifting barrier & 0.884 & 0.363 & 0.211 & 0.213 & 0.156 & 0.156 & 0.102\\
sparse signal & 0.917 & 0.369 & 0.161 & 0.135 & 0.022 & 0.022 & 0.014\\
\bottomrule
\end{tabularx}\end{table}

\section{Appendix Inventory}

The online appendix contains:
\begin{itemize}

\item \texttt{\detokenize{data/raw/summary_trials.csv}}: full trial-level record with 66,150 executed trials;

\item \texttt{\detokenize{data/raw/landscape_metadata.csv}}: 270 landscapes with known optima and signal meta- data;

\item \texttt{\detokenize{data/raw/stability_trials.csv}}: perturbation-stability records;

\item \texttt{\detokenize{data/processed/overall_method_table.csv}}: method-level aggregate table;

\item \texttt{\detokenize{data/processed/main_family_method_table.csv}}: family-method aggregate table;

\item \texttt{\detokenize{data/processed/gated_ep_signal_quality_table.csv}}: signal-quality bin table;

\item \texttt{\detokenize{data/processed/overall_stability_table.csv}}: perturbation stability table;

\item code/: benchmark execution and validation scripts;

\item figures/: diagnostic benchmark figures;

\item SHA256SUMS.txt: integrity checksums.

\end{itemize}
\section{Full Benchmark Protocol}
 The benchmark protocol is deliberately simple enough to audit but rich enough to test the paper's mechanisms. For each landscape family f , landscape index $\ell$, method m, and run index r, the benchmark constructs a landscape Cf,$\ell$ , computes its exact global optimum by enumeration, executes method m under budget T = 45, and records the best candidate returned by the run. The protocol is summarized in Algorithm 2.

\begin{quote}\small
\textbf{Algorithm 2: Full known-optimum benchmark protocol}
\begin{enumerate}[label=\arabic*:,leftmargin=2.4em]
\item For each landscape family, generate 45 landscapes $C_{f,\ell}:\{0,\ldots,100\}\to\mathbb{R}$.
\item Compute $x^\star_{f,\ell}=\arg\min_x C_{f,\ell}(x)$ by exhaustive enumeration.
\item For each of seven methods and each of 35 runs, execute the method with budget $T=45$.
\item Record best visited candidate, best cost, true optimum, global gap, basin indicators, probes, crossings, and signal metadata.
\item Aggregate results by method and family-method condition.
\end{enumerate}
\end{quote}

The full count is therefore

\begin{equation}
6\times45\times7\times35=66{,}150
\end{equation}

executed trials. This count is not a headline approximation; it is the trial count in the raw appendix file.

\subsection{Landscape semantics} The family design isolates specific mechanisms. \texttt{\detokenize{bridge_good}}. The global basin lies beyond the initial basin. A system that cannot cross the boundary is structurally trapped. This family tests the cost of a poor reachable set.

\texttt{\detokenize{bridge_bad}}. The initial basin is already best or near-best. Boundary crossing is unnecessary. This family tests whether escape control can avoid unnecessary exploration.

\texttt{\detokenize{multi_basin}}. Multiple basins compete. The method must either sample broadly or use signals to move between basins.

\texttt{\detokenize{deceptive_no_signal}}. Signals are not trustworthy. This family tests whether controlled search can be worse than broad search when feedback is misleading.

\texttt{\detokenize{sparse_signal}}. Signals are weak until the method is already near informative states. This family tests feedback-density limitations.

\texttt{\detokenize{shifting_barrier}}. Boundary location and value vary across landscapes. This family tests sensitivity to nonstationary reachability structure.

\subsection{Method semantics} The policy set is mechanistic rather than cosmetic. \texttt{\detokenize{reachable_random}}, \texttt{\detokenize{local_repair}}, and \texttt{\detokenize{hard_prune_ep}} deliberately restrict reachability. \texttt{\detokenize{reckless_escape}} deliberately expands reachability without discipline. \texttt{\detokenize{gated_ep_escape}} expands reachability only under a sig- nal threshold. \texttt{\detokenize{oracle_probe_ep}} approximates an upper-bound policy with cleaner probe information. \texttt{\detokenize{random_full}} is the broad-sampling baseline. This design makes the benchmark an identification experiment. If local methods fail on \texttt{\detokenize{bridge_good}}, the cause is reachability. If they succeed on \texttt{\detokenize{bridge_bad}}, the cause is not generic weakness. If gated control improves only with signal quality, the gate signal is the mechanism. If random full search remains competitive under deception, the benchmark rejects a universal-control interpretation.

\section{Proof Details for Control Claims}
\subsection{Detailed proof of the control-value inequality}
 Let J0 be the expected cost-adjusted utility of a baseline policy and Jc the expected cost- adjusted utility of a policy augmented with a control module. Decompose the marginal difference as

\begin{equation}
J_c-J_0=\mathbb{E}[G_P]+\mathbb{E}[G_R]+\mathbb{E}[G_V]-\mathbb{E}[C_D]-\mathbb{E}[C_O]-\mathbb{E}[C_M]
\end{equation}

where the gain terms are improvements from pruning, repair or mutation, and verifier information, while the cost terms are detection cost, overhead, and misdirection. The controlled policy has positive value only when Jc - J0 > 0, which is equivalent to the inequality in the main text. This is a bookkeeping identity, but it is load-bearing: it prevents the paper from treating control complexity as inherently beneficial.

\subsection{Detailed proof of the signal threshold} Let a controlled escape action have two possible epistemic states. With probability p, the signal correctly identifies a beneficial escape and yields benefit B+ . With probability 1 - p, the signal is misleading and yields cost B- . The action also costs CE to probe and execute. The expected value is

\begin{equation}
\mathrm{EV}=pB_+-(1-p)B_- - C_E
\end{equation}

Positive value requires pB+ > (1 - p)B- + CE . Rearranging gives the threshold in the main text. This inequality explains why gated escape is not automatically superior to reckless escape or full random search. A gate helps only when signal reliability exceeds its cost-adjusted threshold.

\subsection{Reachability gap under conservative policies}
Suppose a conservative policy never crosses a barrier, and let $L\subset\mathcal{X}$ denote the initial left basin. If $\mathcal{X}_R\subseteq L$ and the true optimum lies in $\mathcal{X}\setminus L$, then

\begin{equation}
C(x_R^\star)-C(x^\star)\ge\min_{x\in L}C(x)-\min_{y\in\mathcal{X}\setminus L}C(y)
\end{equation}

Thus even perfect optimization inside L cannot close the structural reachability gap. This proposition is exactly instantiated in \texttt{\detokenize{bridge_good}}, where local and hard-pruned methods are trapped.

\paragraph{Proposition 16 (Conservative-trap lower bound).} Let $L\subset\mathcal{X}$ and $\mathcal{X}_R\subseteq L$. If $\min_{y\in\mathcal{X}\setminus L} C(y)<\min_{x\in L} C(x)$, then every reachable optimum has strictly positive globality gap.

\emph{Proof.} Since $\mathcal{X}_R\subseteq L$, $C(x_R^\star)\ge\min_{x\in L} C(x)$. Since the best point outside $L$ is better, $C(x^\star)\le\min_{y\in\mathcal{X}\setminus L} C(y)<\min_{x\in L} C(x)$. Therefore $C(x_R^\star)-C(x^\star)>0$.

\subsection{False-globality risk as a conditional probability} Let H be the event that a method returns a candidate that is good inside its final basin, and let G be the event that the candidate is exact global. False-globality risk can be interpreted as P(H $\cap$ Gc ), with the benchmark implementation operationalizing H through final-basin and reachable-region diagnostics. A high value means that the method is producing plausible local success without exact globality. This is more specific than ordinary error: it is the error mode most likely to support an unjustified global claim.

\section{Additional Reporting Template}

The following template can be used when applying RIO to an AI optimization system.
\begin{enumerate}
\item \textbf{Formal space.} Define the nominal candidate space $\mathcal{X}$, including length, grammar, action horizon, policy class, architecture grammar, or tool-state representation.
\item \textbf{Objective and verifier.} Define $C$ and $V$. State whether the verifier is exact, approximate, stochastic, or proxy-based.

\item \textbf{Generator.} Define the proposal mechanism and initialization distribution $\mu_0$.

\item \textbf{Transition interface.} Define edits, repairs, mutations, tool calls, exploration operators, pruning rules, and memory updates.

\item \textbf{Budget.} State candidate evaluations, verifier calls, tool calls, wall-clock time, context length, or compute budget.

\item \textbf{Reachability evidence.} State what is known about $\mathcal{X}_R$ or $\mathcal{X}_R^\alpha$.

\item \textbf{Optimization evidence.} Report best visited, approximate reachable optimum, or exact global certificate.

\item \textbf{Failure modes.} State expected reachability collapse, overpruning, misdirected escape, verifier gaming, and stability illusions.

\item \textbf{Certificate.} State whether the claim has visited-set, reachable-set, coverage, containment, or complement support.
\end{enumerate}

\textbf{This template is intentionally demanding.} Global optimization language should be earned by the certificate, not by the sophistication of the AI system.

\section{Connection to Practical Learning Pipelines}

\paragraph{Hyperparameter optimization.} The formal space may contain continuous and cat- egorical configurations. In practice, a tuner samples from priors, acquisition functions, early-stopping rules, and resource schedulers. The best configuration is a best visited or reachable optimum unless coverage or regret guarantees apply.
\paragraph{Prompt search.} The formal space of token sequences is vast. Most prompt optimizers use paraphrase templates, mutation rules, retrieved exemplars, or LLM-generated rewrites. The reachable region is determined by those operations. Reporting only the best score hides the rewrite geometry.
\paragraph{Program repair.} The formal space of patches is combinatorial. A repair system reaches patches through edit operators, compiler feedback, tests, and model priors. A patch can be best reachable under visible tests while failing hidden tests or unreachable specifications.
\paragraph{Agent planning.} The formal trajectory space grows exponentially with horizon. The reachable trajectory space is produced by policy priors, tools, action schemas, memory, and environment feedback. A plan can be optimal within the agent's action interface but not globally optimal over all possible plans.
\paragraph{Scientific hypothesis generation.} AI systems that propose hypotheses optimize over a reachable space shaped by literature retrieval, prompt framing, model priors, experiment simulators, and human feedback. The best generated hypothesis is not necessarily the best formal hypothesis in the domain.

\section{Why This Paper Is Not a Leaderboard Claim}
 The benchmark is not designed to declare one universal policy best. It is designed to make a claim distinction measurable. The fact that \texttt{\detokenize{reckless_escape}} performs best in several aggregate metrics is not a recommendation to use reckless escape in deployed systems. It is evidence that reachability expansion matters. The fact that \texttt{\detokenize{random_full}} is strong under poor signal is not evidence against control in general. It is evidence that control without reliable signal can misdirect search. Thus the scientific object is not the method ranking alone. The scientific object is the mechanism map: local policies are stable but trapped, escape expands reachability, gates depend on signal quality, false globality concentrates in trapped methods, and exact-global success requires either broad coverage or reliable reachability expansion.

\section{Online Appendix Integrity}
 The online appendix is part of the evidence object. It contains the raw trial table rather than only selected summaries. The SHA256 manifest allows file-integrity verification. The validation scripts provide a direct path from raw trial records to aggregate claims. This is necessary because the paper's central argument is about claim discipline. A paper about optimization claims should itself make its empirical claim auditable.

\section*{References}
\begingroup\small
\par\hangindent=1.5em\hangafter=1 Bergstra, J. and Bengio, Y. (2012). Random search for hyper-parameter optimization. Journal of Machine Learning Research, 13:281--305.\par
\par\hangindent=1.5em\hangafter=1 Chen, M., Tworek, J., Jun, H., Yuan, Q., Pinto, H. P. de O., Kaplan, J., Edwards, H., Burda, Y., Joseph, N., Brockman, G., et al. (2021). Evaluating large language models trained on code. arXiv:2107.03374.\par
\par\hangindent=1.5em\hangafter=1 Elsken, T., Metzen, J. H., and Hutter, F. (2019). Neural architecture search: A survey.\par
\par\hangindent=1.5em\hangafter=1 Feurer, M., Hutter, F., and others. (2019). Automated machine learning: Methods, systems, challenges. Springer.\par
\par\hangindent=1.5em\hangafter=1 Garnett, R. (2023). Bayesian Optimization. Cambridge University Press.\par
\par\hangindent=1.5em\hangafter=1 Hart, P. E., Nilsson, N. J., and Raphael, B. (1968). A formal basis for the heuristic determi- nation of minimum cost paths. IEEE Transactions on Systems Science and Cybernetics, 4(2):100--107.\par
\par\hangindent=1.5em\hangafter=1 Kocsis, L. and Szepesvari, C. (2006). Bandit based Monte-Carlo planning. In European Conference on Machine Learning.\par
\par\hangindent=1.5em\hangafter=1 Pearl, J. (1984). Heuristics: Intelligent Search Strategies for Computer Problem Solving. Addison-Wesley.\par
\par\hangindent=1.5em\hangafter=1 Russell, S. and Norvig, P. (2021). Artificial Intelligence: A Modern Approach. Fourth edition. Pearson.\par
\par\hangindent=1.5em\hangafter=1 Shahriari, B., Swersky, K., Wang, Z., Adams, R. P., and de Freitas, N. (2016). Taking the human out of the loop: A review of Bayesian optimization. Proceedings of the IEEE, 104(1):148--175.\par
\par\hangindent=1.5em\hangafter=1 Shinn, N., Cassano, F., Gopinath, A., Narasimhan, K., and Yao, S. (2023). Reflexion: Language agents with verbal reinforcement learning. arXiv:2303.11366.\par
\par\hangindent=1.5em\hangafter=1 Silver, D., Schrittwieser, J., Simonyan, K., Antonoglou, I., Huang, A., Guez, A., Hubert, T., Baker, L., Lai, M., Bolton, A., et al. (2017). Mastering the game of Go without human knowledge. Nature, 550:354--359.\par
\par\hangindent=1.5em\hangafter=1 Snoek, J., Larochelle, H., and Adams, R. P. (2012). Practical Bayesian optimization of machine learning algorithms. In Advances in Neural Information Processing Systems.\par
\par\hangindent=1.5em\hangafter=1 Sutton, R. S. and Barto, A. G. (2018). Reinforcement Learning: An Introduction. Second edition. MIT Press.\par
\par\hangindent=1.5em\hangafter=1 Yao, S., Zhao, J., Yu, D., Du, N., Shafran, I., Narasimhan, K., and Cao, Y. (2023). ReAct: Synergizing reasoning and acting in language models. In International Conference on Learning Representations.\par
\endgroup

\end{document}